\documentclass[10pt,twocolumn,letterpaper]{article}

\usepackage{iccv}
\usepackage{times}
\usepackage{epsfig}
\usepackage{graphicx}
\usepackage{amsmath}
\usepackage{amssymb}
\usepackage{booktabs}
\usepackage{algorithm}
\usepackage{algorithmic}

\usepackage{multirow}
\usepackage{caption}
\usepackage{color}
\usepackage[toc,page]{appendix}
\newcommand{\ve}[1]{\mathbf{#1}} 

\iccvfinalcopy 

\def\iccvPaperID{****} 

\ificcvfinal\fi

\begin{document}

\title{Two-Stream Video Classification with Cross-Modality Attention}

\author{Lu Chi, Guiyu Tian, Yadong Mu$^\ast$\\
Peking University\\
{\tt\small \{chilu,wavey,myd\}@pku.edu.cn}
\and
Qi Tian\\
Huawei Noah's Ark Lab\\
{\tt\small tian.qi1@huawei.com}
}

\maketitle
\ificcvfinal\thispagestyle{empty}\fi

\begin{abstract}
  Fusing multi-modality information is known to be able to effectively bring significant improvement in video classification. However, the most popular method up to now is still simply fusing each stream's prediction scores at the last stage. A valid question is whether there exists a more effective method to fuse information cross modality. With the development of attention mechanism in natural language processing, there emerge many successful applications of attention in the field of computer vision. In this paper, we propose a cross-modality attention operation, which can obtain information from other modality in a more effective way than two-stream. Correspondingly we implement a compatible block named CMA block, which is a wrapper of our proposed attention operation. CMA can be plugged into many existing architectures. In the experiments, we comprehensively compare our method with two-stream and non-local models widely used in video classification. All experiments clearly demonstrate strong performance superiority by our proposed method. We also analyze the advantages of the CMA block by visualizing the attention map, which intuitively shows how the block helps the final prediction.
\end{abstract}

\section{Introduction}

\renewcommand*{\thefootnote}{\fnsymbol{footnote}}
\footnotetext{$^\ast$ is the corresponding author.}
\setcounter{footnote}{0}
\renewcommand*{\thefootnote}{\arabic{footnote}}

In recent years, thanks to the emergence of massive video datasets~\cite{Abu-El-HaijaKLN16,KayCSZHVVGBNSZ17Kinetics}, applications of deep learning in video classification have witnessed a rapid development. However, there is still a considerable improvement space towards human-level video understanding. The state-of-the-art video classification methods are mainly based on convolutional neural networks. Despite tremendous progress has been recently made in designing highly discriminative network architectures, there still remain many open research problems. This research essentially concerns the following two problems:

Firstly, an insufficiently explore problem in video understanding is a more powerful way to capture the dynamic information or motions in videos. As one of the main differentiators between videos and images, dynamic information is regarded to be indispensable for effective video classification. For example, it is a difficult task even for a human being to discriminate from different kinds of dances (\emph{e.g.}, \textit{salsa dancing}, \textit{tango dancing} and \textit{dancing macarena}) by only having a glimpse at a single frame. A large number of widely-known video semantic categories, such as dancing and other sports, can be faithfully classified when we can extract sufficient motion-related information like moving trajectories.

Secondly, subtle details are key for recognizing some video categories or actions. The literature still lacks some in-depth analysis on effectively attending to those discriminative video details. Attention plays an important role in the field of natural language processing and image recognition. But it is still a nascent research topic in video action recognition. By grasping subtle details, humans can easily distinguish many classes. Considering the action \textit{sword fighting} or \textit{playing cricket}, only a single frame is enough as long as you can find a sword or a cricket. Generally, video motions attract more human attention and are likely to be related to key clues. For example, for the two actions \textit{making a sandwich} and \textit{making pizza}, their key objects sandwich or pizza are both around the moving hands. In this situation, motion can help attention.

Motivated by these observations, we propose a cross-modality attention operation, which can make full use of both static and motion information. Unlike the classic two-stream framework~\cite{simonyan14two} that fuses information from two modalities in a late stage, we fuse the information in a more hierarchical and effective way.

Our proposed cross-modality attention operation devises such an attention mechanism that it encourages one modality absorbs most complementary information from other modalities.
In contrast to the recently-proposed non-local operation~\cite{NonLocal2018}, the proposed cross-modality attention can pay attention to other modalities rather than being constrained in the same modality. When two modalities under investigation are identical, our proposed method boils down to the non-local operation. Another key trait is that attentions are computed in a global manner. Specifically, spiritually alike the non-local operation, our proposed method computes the response as a weighted sum of the features of the other modality at all positions.

There are three main advantages of using the proposed cross-modality attention operation, sketched briefly as below: 1) It can effectively fuse the information between two or more modalities; 2) It can capture long-range dependencies by globally investigating the feature maps; 3) It can be wrapped as a highly compatible block that can be inserted into almost all existing neural networks and frameworks.

The rest of this paper is organized as: we first review related work in Section 2 and detail the novel cross-modality attention operation / network design in Sections 3 and 4. Section 5 shows the experiments and detailed analysis of our module.

\section{Related Work}

With the significant development of deep learning in image recognition~\cite{Krizhevsky2012ImageNet,SimonyanZ14a,HeZRS16ResNet,HuangLMW17dense}, a large number of active researches have emerged in the field of video classification. Karpathy \emph{et al}.~\cite{KarpathyTSLSF14} contributed significant breakthrough in the video classification task. Their major contribution is 3-D convolutional neural networks trained on the Sports-1M data, which far exceeds traditional methods~\cite{WangUKLS09eval,NieblesCF10} in terms of top-1 or top-5 classification accuracies. This seminal work demonstrates the power of temporal information in video-related tasks.

Optical flow fields are known as conceptually simple and empirically effective when attempting to capture the temporal information. A variety of approaches have been developed to utilize optical flow in video classification. A large body of existing works~\cite{simonyan14two,Wu2015modeling,CarreiraZ17i3d,sun2018optical} has re-iteratively found that feeding the optical flow fields into a deep model can bring comparable performance with the RGB stream in video classification. After properly fused via late-stage fusion, one can accomplish a performance better than either stream. Recent endeavor along this research thrust includes direct mapping two adjacent frames to the optical flow field~\cite{MayerIHFCDB16,IlgMSKDB17flownet}. Researchers have also investigated using deep neural networks for computing optical flows, which can be expedited by modern GPU hardware. However, the major obstacle stems from the lack of high-quality training data. To mitigate the data scarcity, some train an optical flow model from synthesized datasets~\cite{MayerIHFCDB16}, or predict the label of videos in an end-to-end way for improving the accuracy\cite{Lara2017integration,NgCND18actionflownet}. In addition, the optimization ideas in the traditional methods are integrated into the design of the neural networks. Fan \emph{et al}.~\cite{fan2018end} unfold the optimization iterations in TV-L1~\cite{ZachPB07tvl1} as neural layers, and Sun \emph{et al}.~\cite{sun2018optical} propose neural networks to learn the representations orthogonal to the optical flow. We would point out that, though tremendous efforts have been noticed in computing optical flows, litter has been done to explore how to effectively using optical flow in video classification.

Optical flow can be regarded as an explicit way to utilize motion information to video classification. More recent research is pursuing other alternatives that rely on deep neural networks is automatically distill spatio-temporal video information. Typical examples include inflating 2D convolution into 3D convolution~\cite{JiXYY13,TranBFTP15C3D,tran2018closer}. One of key weaknesses of these models are the gigantic parameters used for defining high-dimensional convolutions etc. Using pre-trained models is a popularly-verified effective strategy for easing the model training in many tasks~\cite{RenHG017,Chen2017CPN,Jacob2018BERT}, such as transferring deep models pre-trained on ImageNet to 3D CNN. A naive solution is to duplicate the parameters of the 2D filters $T$ times along the time dimension, and rescale all parameters by dividing by $T$~\cite{CarreiraZ17i3d}. This ensures a same response from the convolution filters. To reduce parameter number in 3D CNNs, some works factorize 3D convolutional filters into separate spatial and temporal components and strike a compromise in accuracy and efficacy~\cite{qiu2017P3D,tran2018closer,xie2017rethinking}. Other relevant works mix 3D convolution with 2D convolution in a neural network~\cite{wang2017arnet,xie2017rethinking,Zolf2018ECO}. Despite the empirical success on indicative video benchmarks, 3D CNNs are far from reaching the point of fully acquiring the motion information and replacing the optical flow. Fusing with the optical flow stream is still an effective practice~\cite{CarreiraZ17i3d,wang2017arnet}. In fact, we can regard 3D CNNs as a general tool that acquires relation among adjacent frames. It can be fed with either RGB frames or other modalities (\eg, optical flow).

For complex video objects, other information also provide complementary information, including audio~\cite{long2018attention}, human pose~\cite{du2017rpan,liu2018recognizing}, and semantic body part~\cite{Zhao2017body} etc. Learning how to efficiently integrate multi-modality information is an emerging research direction. Existing researches, such as pooling at different stages~\cite{KarpathyTSLSF14,NgHVVMT15Beyond} or modeling long-range temporal structure using LSTM~\cite{Wu2015modeling}, mainly concern fusing in the temporal dimension. There are rarely relevant studies about the fusion of different modalities~\cite{Wu2015modeling}. To date, the mainstream method is still the two-stream method~\cite{simonyan14two}. Our primary goal is to design a network structure that is more effective than two-stream and meanwhile achieves higher precision.

Attention networks have been originally popularized in natural language processing~\cite{BahdanauCB14,VaswaniSPUJGKP17attention}, used for comprehension and abstractive summarization etc. Recent years have observed a quick spread in computer vision~\cite{SharmaKS15,Fu18dual,abs-1805-08318}. Xie \emph{et al}~\cite{xie2017rethinking} place a feature gating module after some convolutional layers to weight the features in each channel in an adaptive, data-dependent way. Long \emph{et al}~\cite{long2018attention} propose attention clusters, which aggregates local features to generate a valid global representations for video classification. Non-local networks~\cite{NonLocal2018} can weight all information (including space and time) by adopting a mechanism similar to self-attention. Our motivating observation is the lack of cross modality attention block, which works globally as non-local block but can make full use of cross-modality information. Importantly, this block shall be compatibly inserted into most existing network structures including the classic two-stream inputs.

\section{Cross Modality Attention}

In this section, we give detailed description of the proposed Cross Modality Attention(CMA) operation and its implementation.

\subsection{Formulation}

Our proposed Cross Modality Attention(CMA) operation can be precisely described in the Q-K-V language, namely matching a query from one modality with a set of key-value pairs from the other modality and thereby extracting most critical cross-modality information. Following the notations in~\cite{VaswaniSPUJGKP17attention}, we define the generic CMA operation as:
\begin{equation}
    \text{CMA}(Q_{1},K_{2},V_{2})=softmax \left(\frac{Q_{1}K_{2}^{T}}{\sqrt{d_{k}}} \right) V_{2},
\end{equation}
where the index $1$ or $2$ represents different modality. $Q$ is the set of queries, $K$ is a matrix of the memory keys and $V$ contains memory values. All Q, K and V are of feature dimension $d_{k}$.

Here we give a concrete instance of the CMA operation in neural networks. Given a typical two-stream data (RGB + flow), a CMA operation can be written as:
\begin{eqnarray}
\ve{z}_{i} &=& \frac{1}{\mathcal{C}(\ve{x},\ve{y})}\sum_{\forall{j}}f(\ve{x}_{i},\ve{y}_{j})v(\ve{y}_{j}) \label{equ:cma-operation} \\
f(\ve{x}_{i},\ve{y}_{j}) &=& e^{q(\ve{x}_{i})k(\ve{y}_{j})^{T}} / \sqrt{d_{k}} \\
\mathcal{C}(\ve{x},\ve{y}) &=& \sum_{\forall{j}}f(\ve{x}_{i},\ve{y}_{j}),
\end{eqnarray}
where $\ve{x}$ is from the feature maps of specific stage of the RGB branch, such as the output of $\text{res}_{4}$ in ResNet~\cite{HeZRS16ResNet}. $\ve{y}$ is from the feature maps in the flow branch. $\ve{z}_{i}$ denotes the output of the CMA operation. $i$ and $j$ are both indices of feature maps (can be in space, time, or spacetime). $q$, $k$ and $v$ are linear embeddings which map $\ve{x}$ or $\ve{y}$ to queries, keys and values of $d_{k}$ dimensions respectively. The function $f$ can be flexibly defined, with many instantiations discussed in~\cite{NonLocal2018}. For simplicity, we choose the embedded Gaussian version in this paper.

The non-local operation~\cite{NonLocal2018} is essentially self-attention and only pays attention to intra-modality. In comparison, our proposed CMA is cross-modal. Moreover, the non-local operation can be regarded as a special case of CMA when $K$, $Q$ and $V$ are all from the same modality.

\subsection{CMA Block}
\label{sec:cma}

A CMA block is a wrapper of the CMA operation that can be inserted into many existing neural networks, which is defined as:
\begin{equation}
out_{i}=W_{out}\ve{z}_{i}+\ve{x}_{i},
\end{equation}
where $\ve{x}_{i}$ and $\ve{z}_{i}$ are given in Eqn.~(\ref{equ:cma-operation}). $W_{out}$ defines a linear embedding that can be implemented by convolution operation.

\begin{figure}[t]
\centering
\includegraphics[width=.95\linewidth]{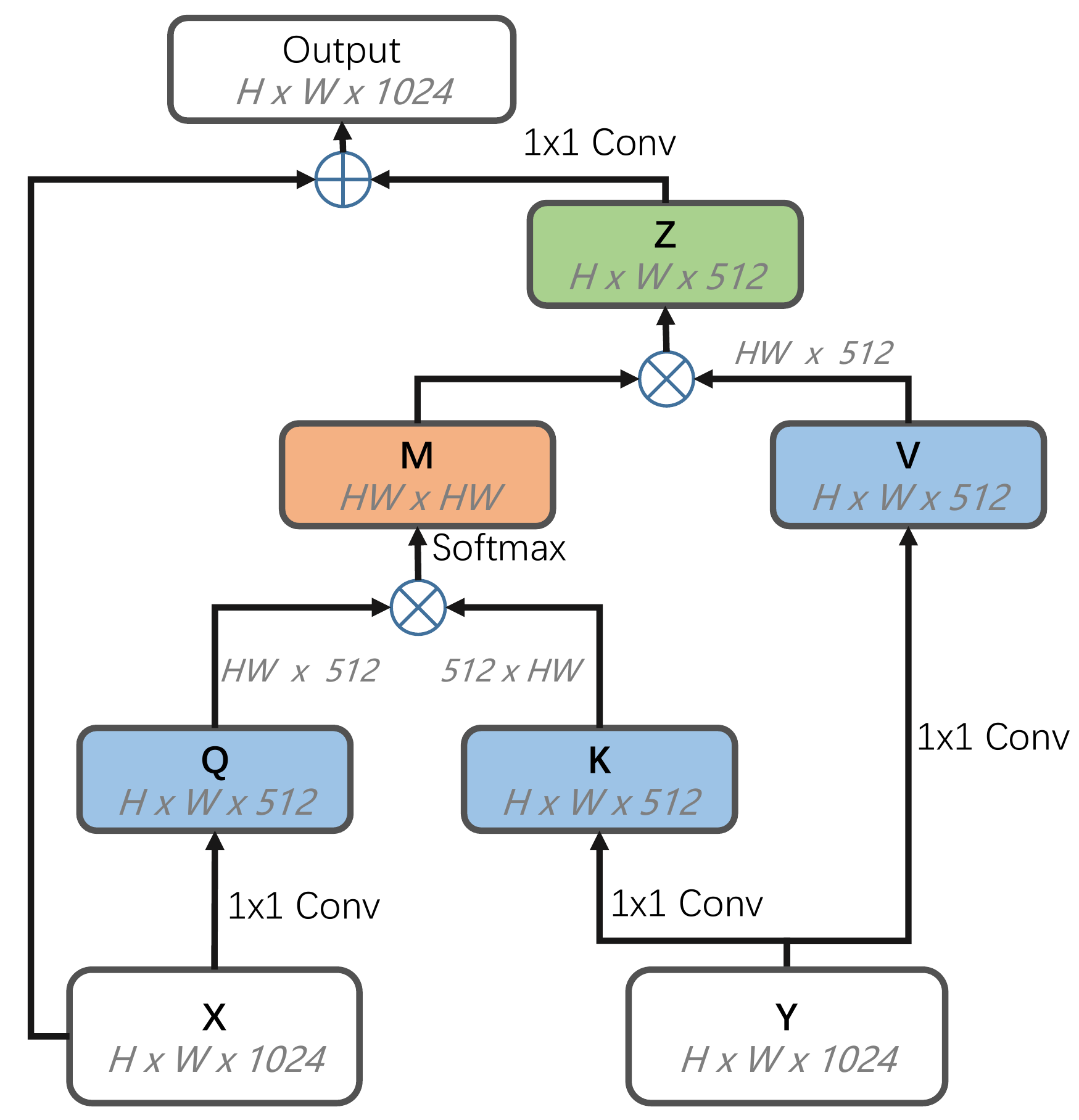}
\caption{\small An example of \textbf{CMA block}. We show the shape of feature maps at each stage, such as $H\times W\times 1024$, where $1024$ is the number of channels. Let $X$ be the feature maps of the RGB branch and $Y$ be the feature maps of flow branch. The number of channels is halved via $1\times1$ convolutions. Reshaping or transposing is performed whenever needed. ``$\bigotimes$" denotes matrix multiplication, and ``$\bigoplus$" denotes element-wise sum.}
\label{fig:block}
\end{figure}

Figure ~\ref{fig:block} presents an example of the CMA block, where $Q$ comes from the RGB branch and $V$, $K$ come from the flow branch. This allows the RGB branch to attend over all positions in the flow branch at a specific stage. As a result, it can get more valuable information selectively from the flow branch which may be weak or even missing in itself.
A CMA block can be added into any location of deep neural networks, since it can be fed with input of any shape and ensure a same-shaped output. This flexibility allows us to fuse richer hierarchical features between different modalities. To make the CMA block more compatible, we add a residual connection ``$+\ve{x}_i$"~\cite{HeZRS16ResNet}. This guarantees a non-worse accuracy with the CMA block by some simple means (\emph{e.g.}, zeroing $W_{out}$).

\begin{figure*}[t]
\centering
\includegraphics[width=.95\linewidth]{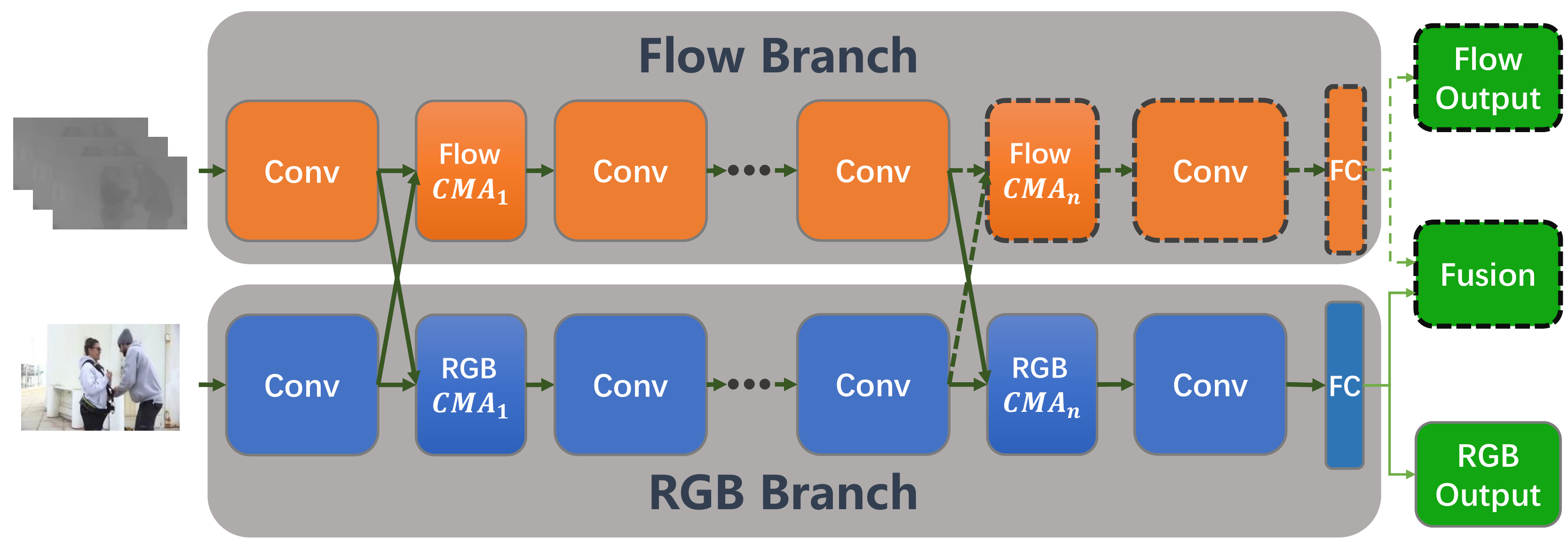}
\caption{\small \textbf{An overview of the video classification model.} This model contains both RGB branch and Flow branch. In each branch, we insert $n$ CMA blocks, which play an important role in transmitting information between different branches. There are three outputs in this model. Operations in dotted box are not essential since practically we can only use the output of the RGB branch for prediction. }
\label{fig:architecture}
\vspace{-0.1in}
\end{figure*}

\textbf{Implementation of CMA Blocks}: We implement functions $q$, $k$, and $v$ as $1\times1$ convolutions in space or $1\times1\times1$ convolution in space-time, denoted as $\text{conv}_{q}$, $\text{conv}_{k}$ and $\text{conv}_{v}$ respectively. To reduce the computational complexity and GPU memory consumption, we let $\text{conv}_{q}$ reduce the number of channels to be half of that in $\ve{x}$, and $\text{conv}_{k}$ and $\text{conv}_{v}$ have same number of channels as $\text{conv}_{q}$. Note that we also insert a spatial max-pooling layer with stride 2 before $k$ and $v$, further simplifying the computation. Inspired by~\cite{NonLocal2018}, a batch normalization layer is added after $W_{out}$ and has the scale parameters initialized to be zeros, which sets the entire CMA block as an identity mapping.

\section{Network Architecture}

This section elaborates on our proposed video classification model. We first introduce two branches in our model and how the CMA blocks are inserted. Then, we depict the whole network architecture and finally describe the details of training strategy and implementation.

\subsection{Two Branches for video classification}
\label{sec:architecture_branches}

As shown in Figure~\ref{fig:architecture}, our model contains two branches, including the RGB branch and Flow branch. As mentioned in~\cite{simonyan14two}, the RGB branch carries visual information about scenes and objects in the video, while the Flow branch conveys the motion. All the information from both branches are crucial for video classification. In two-stream~\cite{simonyan14two}, they simply average the scores of the two branches to make the final prediction.

We add several CMA blocks at some intermediate stages of each branch, obtaining information from the other branch. Compared with the two-stream method, this fuses two modalities much earlier and hierarchically. There are three classification scores in our model. The first two scores are from the RGB branch and the Flow branch respectively, and the last one is a weighted summation of RGB / Flow branches. Empirical investigation deferred to Section~\ref{sec:experiment} shows that any of these three scores can make an excellent prediction. In fact, in the scenario of highly-budgeted parameters, one can just use the scores of the RGB branch without much loss of performance.

\textbf{Implementation}: Both branches adopt ResNet-50~\cite{HeZRS16ResNet} as base network. Considering the limited GPU memory and precise spatial information, we add 5 CMA blocks in $\text{res}_{3}$ and $\text{res}_{4}$ to every other residual block, which is also similar to the setting in non-local neural networks~\cite{NonLocal2018}. The RGB branch takes only one RGB frame as input, while the Flow branch stacks five consecutive optical flow fields as input. The RGB branch can be directly initialized from the ResNet weights pre-trained on ImageNet~\cite{Olga2015ImageNet}. Since the number of input channels of the Flow branch is different from that of the models pre-trained on ImageNet, we initialize the weights of the first convolution by replicating the means of the pre-trained weights across channels. The CMA blocks are initialized via the same scheme in~\cite{HeZRS16ResNet}. We zero the scale parameters of the last BN layer as previously mentioned in Section~\ref{sec:cma}.

\subsection{TSN Framework}
Temporal Segment Networks (TSN) has been proved to be powerful in modeling long-range temporal structure~\cite{wang16tsn,wang2017arnet,sun2018optical}. We also incorporate this effective albeit simple framework. Given a video, we divide it into $K$ segments, ensuring the duration of each segment equal. For each segment, a snippet (1 RGB frame for the RGB branch and 5 consecutive optical flow fields for the Flow branch) is randomly sampled. We average the scores produced by each segment to get the final video-level score, namely
\begin{equation}
    \ve{G}=\frac{1}{K}\sum_i^K\ve{G_i},
\end{equation}
where $K$ is the number of segments and $\ve{G_i}$ is the score of one specific snippet.

The overall loss function can be defined as:
\begin{equation}
    \mathcal{L}(y,\ve{G})=-\sum_{c=1}^{C}y_{c}(G_c-\log\sum_{j=1}^{C}e^{G_j}),
\end{equation}
where $C$ is the number of video classes and $y_c$ is the ground-truth label concerning class $c$. $G_c$ are the scores of the same class on all snippets.

\subsection{Training Strategy}
\label{sec:train-strategy}

Since the Flow models converge much slowly than RGB models~\cite{wang16tsn}, we firstly train the flow branch on Kinetics data~\cite{KayCSZHVVGBNSZ17Kinetics}. After that, considering the limited GPU memory, we train the CMA Model in an iterative way between two branches. Thinking that the CMA blocks is initially an identity mapping and the Flow branch has been trained on the kinetics, the Flow branch can provide more reliable information to the RGB branch before the iterative training stage. Therefore, we train the RGB branch in $iter_1,iter_3,iter_5...$ and train the Flow branch in $iter_2,iter_4,iter_6,...$. When training the RGB branch, its parameters are optimized according to the loss of the current branch and we freeze all the layers in the Flow branch, including CMA blocks of the Flow branch. Similar treatment for training the Flow branch. The total number of epochs at each iteration is set to 30.

\subsection{Implementation Details}

\textbf{Input}: The video frames are scaled to size $256 \times 256$. We choose the TVL1 optical flow algorithm~\cite{ZachPB07tvl1} to extract optical flow for the Flow branch, based on the GPU version from the OpenCV toolbox. The pixel values of optical flow are truncated to the range $[-20,20]$, and then re-scaled between -1 and 1. The input size of two branches are both $224 \times 224$, cropped from the video frames or optical flow fields. The RGB branch takes only one frame ($frame_{t}$) as the input and the Flow branch reads a stack of consecutive optical flow fields ([$of_{t}$,$of_{t+1}$,$of_{t+2}$,$of_{t+3}$,$of_{t+4}$]). In other words, the input shapes of two branches are $N\times224\times224\times3$ and $N\times224\times224\times10$ respectively, where $N$ is the batch size and the last dimension represents the number of channels. It's important to note that the RGB frame is corresponding to the first optical flow field in the temporal dimension, and all RGB frame / optical flows are spatially aligned. For data augmentation, we use random horizontal flipping, random cropping and scale jittering~\cite{wang16tsn}. And the number of segments is set to 3.

\textbf{Training}: We use a standard cross-entropy loss and mini-batch stochastic gradient descent algorithm to optimize the network parameters, where the batch size is 128. We train the model with BN enabled, which is the same to~\cite{NonLocal2018}. To make the statistics of each BN layer more accurate, we use the synchronized batch normalization~\cite{TeTeXiao17MegDet}. The learning rate is initialized as 0.01 and get reduced by a factor of 10 when the accuracy is stuck in some plateau. At the beginning of each iteration, we reset the learning rate to the initial value. The dropout ratio is 0.7 and the weight decay is $5 \time 10^{-4}$, which are introduced to reduce over-fitting.

\textbf{Testing}: During test time we use ten-croppings and flip four corners and the center of the frame or optical flow filed as~\cite{Krizhevsky2012ImageNet}. The number of segments is set to 25 and the temporal spacing between each segment is equal. We average the scores across all the samples and crops of them to get the final video-level score. For the fusion score, we firstly get the frame-level scores via weighted sum and then average all the scores to get the video-level score. We will provide an empirical study on the fusion weights in Section~\ref{sec:fusionweight}.

\section{Experiment}
\label{sec:experiment}

We evaluate the proposed methods and perform ablation studies on two popular datasets, \textbf{UCF101}~\cite{Soomro12ucf101} and \textbf{Kinetics}~\cite{KayCSZHVVGBNSZ17Kinetics}. For clarity, let CMA\_$iter_i$ be the model trained after $i$th iterations. We add the suffix ``-R", ``-F", ``-S" for the RGB / Flow streams or two-stream respectively.

\subsection{Dataset}

\textbf{UCF-101}~\cite{Soomro12ucf101} consists of 101 action classes and over 13-K clips. All the videos are downloaded from YouTube, and all of them are recorded in unconstrained environments, including various lighting conditions, partial occlusion, low quality frames etc.

\textbf{Kinetics}~\cite{KayCSZHVVGBNSZ17Kinetics} is a large-scale trimmed video dataset which contains more than 300-K video clips in total, and each clip has a duration of around 10 seconds. The dataset covers 400 human-centric classes and each class has at least 400 video clips. For unknown reasons, there are some invalid urls and we are unable to crawl some of the videos. We get 232,679 videos for training and 19,169 for validation. We skip processing the testing set since their labels are not provided.

\subsection{Investigation of Fusion Weights}
\label{sec:fusionweight}

\begin{figure}[t]
\centering
\includegraphics[width=.95\linewidth]{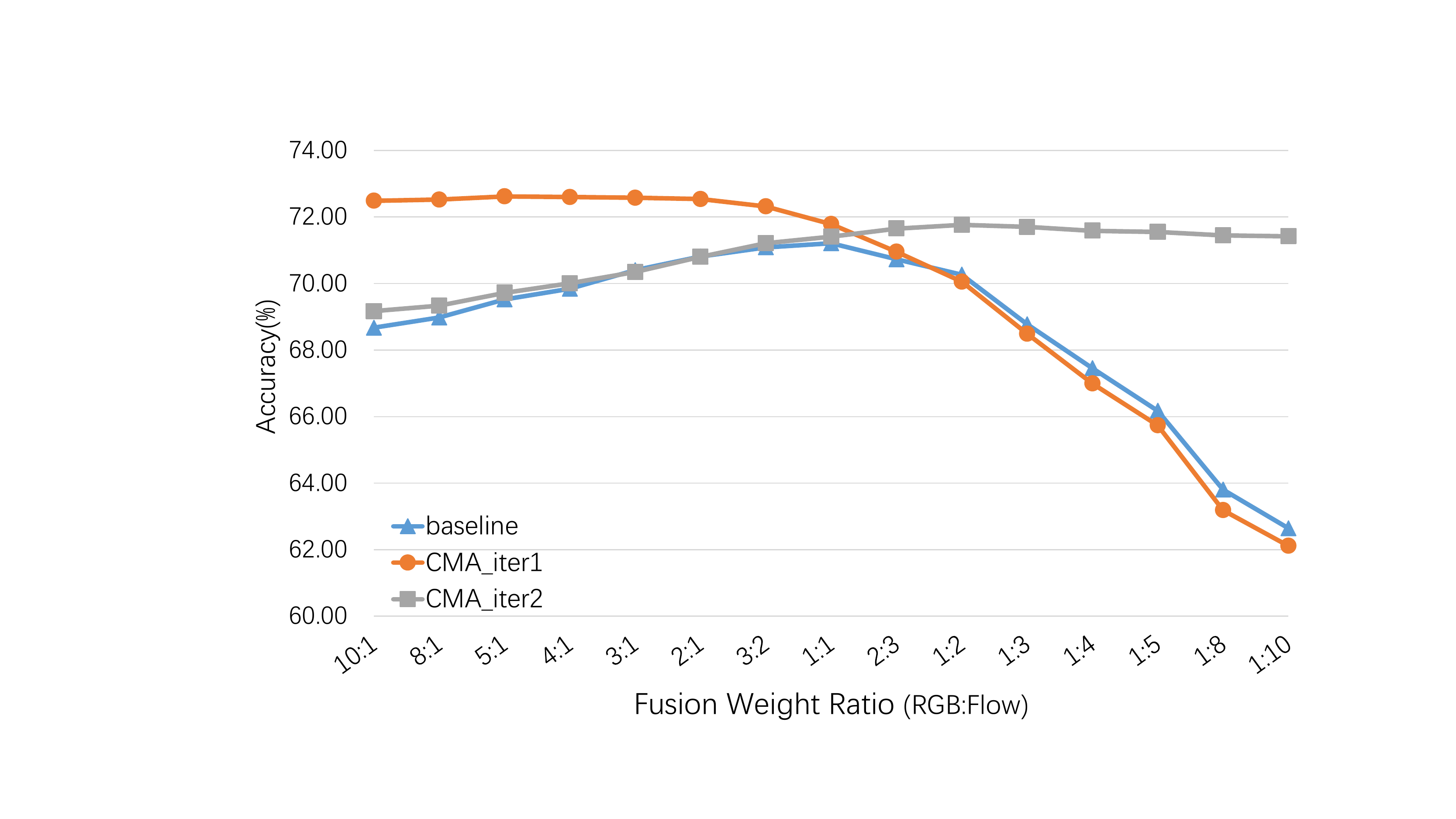}
\caption{\small \textbf{The top-1 accuracies with different fusion weights.} The CMA models perform better than the two-stream when we give higher weight to the more reliable branch.}
\label{fig:fusion}
\centering
\end{figure}

To get the fusion score, two-stream~\cite{simonyan14two} averages the scores from two modalities and \cite{wang16tsn} gives more credits to the RGB modality by setting its weight as 1 and that of Flow modality as 1.5. But our proposed model contains two branches which are interdependent, consequently, training one branch inevitably have an effect on the other one. In this situation, exploring suitable fusion weights is necessary. Figure~\ref{fig:fusion} shows the top-1 accuracy with different fusion weights. We use the two-stream as a baseline whose base model is the same as ours (ResNet50). The two-stream model can achieve a higher accuracy when the weight of the two branches is almost the same. But for the CMA models, at the first training iteration, we just train the RGB branch with the Flow branch fixed, so the RGB branch performs better than the Flow branch. In other words, the RGB branch is more reliable. Giving the RGB branch more weight will make the final accuracy higher, but too much weight will make the other branch almost completely ignored. At the second iteration, we should similarly give more weight to the Flow branch. From Figure~\ref{fig:fusion} one can see that the fusion accuracy of CMA models is always higher than the baseline, as long as we give more weight to the more reliable branch.

Based on the above analysis, we give the equal weights to the two branches in two-stream, identical to~\cite{simonyan14two}, and set the weights of the RGB / Flow branches as $5:1$ at $iter_1,iter_3,...$ and $1:5$ at $iter_2,iter_4,...$. For all the following experiments we adopt such setting.

\subsection{Performance at Each Iteration}

\begin{table}[thb]
\centering
\begin{small}
\begin{tabular}{ccccccc}
\toprule
\multirow{2}{*}{Iteration} & \multicolumn{2}{c}{RGB} & \multicolumn{2}{c}{Flow} & \multicolumn{2}{c}{two-stream} \\
\cmidrule(lr){2-3} \cmidrule(lr){4-5} \cmidrule(l){6-7}
  & top-1 & top-5 & top-1 & top-5 & top-1 & top-5 \\
\cmidrule(r){1-3} \cmidrule(lr){4-5} \cmidrule(l){6-7}
0 & 67.73 & 87.94 & 55.73 & 79.04 & 71.21 & 89.92 \\
1 & 72.17 & \textbf{90.70} & 55.73 & 79.04 & \textbf{72.62} & \textbf{91.04} \\
2 & 68.45 & 88.54 & \textbf{71.17} & \textbf{90.12} & 71.55 & 90.24 \\
3 & \textbf{72.19} & 90.63 & 69.81 & 89.41 & 72.55 & 90.82 \\
\bottomrule
\end{tabular}
\end{small}
\caption{\small Accuracies at each iteration on the Kinetics dataset.}
\label{tab:iter}
\end{table}

In Section~\ref{sec:train-strategy}, we introduce the iterative training strategy. Here let us study how many iterations we need for convergence. Table~\ref{tab:iter} lists the accuracy at different iterations. $iter_0$ represents the baseline that has the two branches trained independently. The fusion accuracy is equal to two-stream~\footnote{Although we name the baseline as $iter_0$, we don't initialize the CMA model with the parameters in $iter_0$. The train strategy keep the same as described in Section~\ref{sec:train-strategy}}. After $iter_1$, the RGB branch has exceeded the two-stream, and the Flow branch keeps the same as the baseline because we have not trained on it at this iteration and the CMA blocks in this branch are now just an identity mapping. Additionally, the accuracy of fusion is much higher than others. In order to achieve higher accuracy for the Flow branch, we train the Flow branch with the RGB branch freezed at the second iteration. As expected, the accuracy of the Flow branch is improved and can be comparable to the two-stream. But the performance of the RGB branch drops due to that the distribution of the feature maps of the Flow branch has changed, which can affect the RGB branch through the CMA blocks. After $iter_3$, the accuracy of the RGB branch returns to the relative high level while the Flow branch degrades slightly. The fusion score doesn't be improved any more. It is thus observed that the first iteration is almost sufficient for our models.

\subsection{Analysis and Visualization}

\begin{table}[thb]
\centering
\begin{small}
\begin{tabular}{cccc}
\toprule
model & params & top-1 & top-5 \\
\cmidrule(){1-4}
ResNet50-R & 1$\times$ & 67.73 & 87.94 \\
two-stream & 2$\times$ & 71.21 & 89.92 \\
CMA\_$iter_1$-R & \textbf{1.8}$\times$ & \textbf{72.17} & \textbf{90.70} \\
\bottomrule
\end{tabular}
\end{small}
\caption{\small CMA model \emph{vs} two-stream in terms of parameter number and accuracy. The number of parameters are relative to the ResNet50 baseline.}
\label{tab:vstwostream}
\end{table}

Table~\ref{tab:vstwostream} compares our method with two-stream in terms of a few key factors, including number of parameters and final accuracy. CMA\_$iter_1$-R is more accurate than two-stream, though fewer parameters are used. That validates that our CMA model is more effective than two-stream for fusing.

\begin{figure*}[t]
\centering
\includegraphics[width=0.98\linewidth]{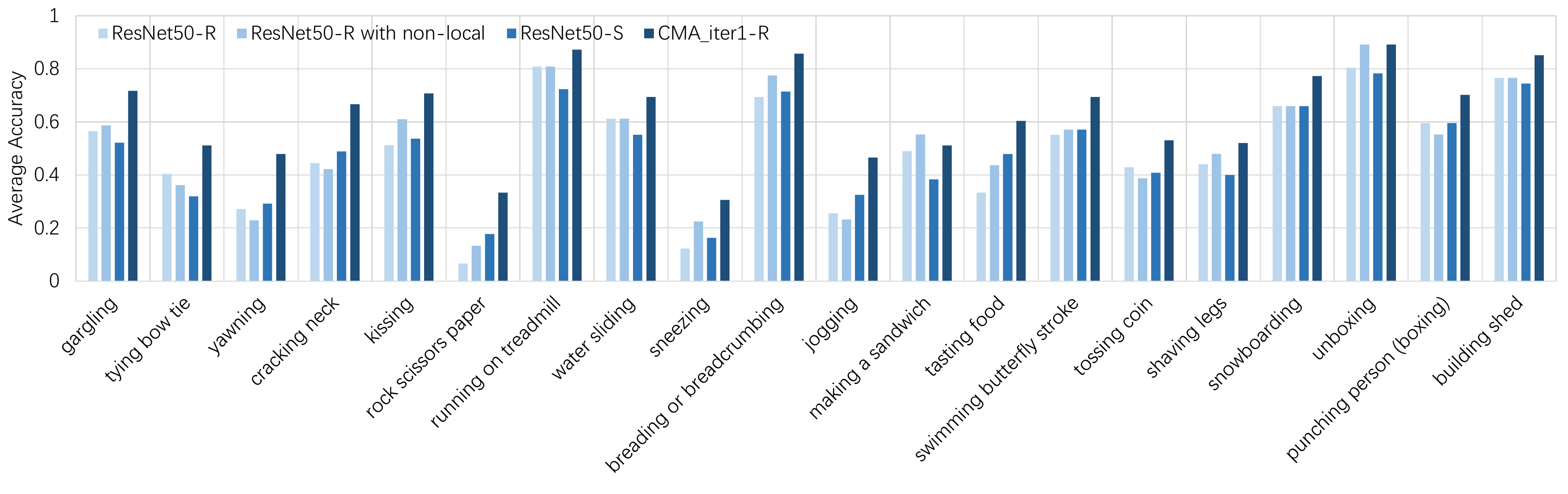}
\caption{\small \textbf{Comparing top-20 most improved categories between the proposed CMA model and two-stream}.}
\label{fig:cma-two-stream}
\centering
\end{figure*}

\begin{table}[thb]
\centering
\begin{small}
\begin{tabular}{cc}
\toprule
groundtruth & confusing category \\
\cmidrule(r){1-1} \cmidrule(l){2-2}
gargling & trimming or shaving beard \\
tying tie & tying bow tie \\
yawning & baby waking up \\
cracking neck & massaging back \\
kissing & hugging \\
rock scissors paper & shaking hands \\
running on treadmill & waxing leg \\
water sliding & jumping into pool \\
sneezing & crying \\
breading or breadcrumbing & cooking chicken \\
\bottomrule
\end{tabular}
\end{small}
\caption{\small The top-10 confusing categories on which the CMA model achieves the largest gain compared with two-stream in Kinetics. The gain is the improved accuracy (\%).}
\label{tab:confusing}
\end{table}

Figure~\ref{fig:cma-two-stream} showed the top-20 most improved categories and compare between our CMA model / two-stream. We also list the top-10 confusing categories in Table~\ref{tab:confusing}. Compared with two-stream, the proposed CMA model is more sensitive about the motion trajectories, such as \textit{water sliding} and \textit{jumping into pool}, although the background is similar. Due to the fact that almost all of the samples of \textit{yawning} are about babies, it's very easy to confuse with \textit{baby waking up}. But our model can improve the performance according to the different motion between two categories. Additionally, the CMA model can pay more attention to the moving objects, such as jaw or mouth, or the tools (cup or razor) held on the hand, while discriminating between \textit{gargling} and \textit{trimming or shaving beard}. We also visualize some attention maps in Figure~\ref{fig:attentionmap}.

\begin{figure*}[thb]
\centering
\includegraphics[width=.95\linewidth]{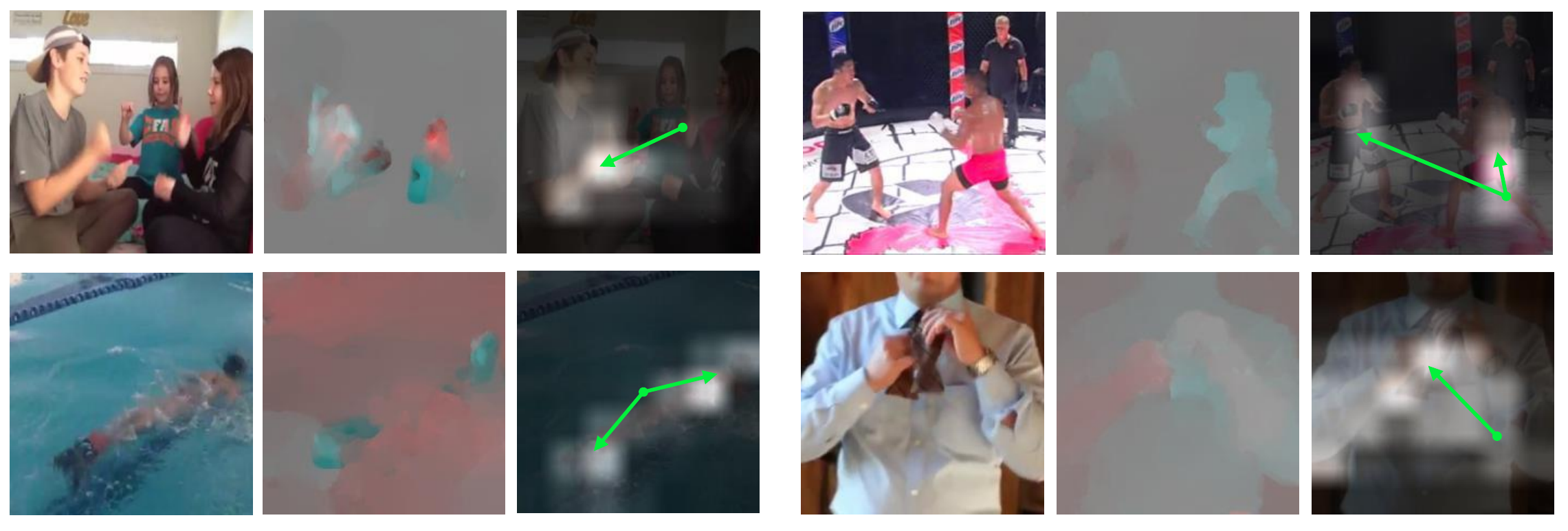}
\caption{\small \textbf{Examples of the attention maps}. We train CMA\_$iter_1$-R on Kinetics and visualize the attention maps of the last CMA block in res$_4$ since the last block is the most related to the final classification. These samples are taken from Kinetics randomly. Each set contains three images, including (from left to right) RGB frame, optical flow fields, and the attention map. In the attention map, we draw some arrows that start from the query location and point to the more interesting parts in the CMA block. We observe that the block can easily focus on moving objects, such as the moving hand in the top-left set and the swimming person in the bottom-left set. And as a result, the RGB branch can take important information from the Flow branch as much as possible within limited capacity. We also find that the CMA operation is global. Looking at the example in the top-right, the pixel on the right person can not only focus on the nearby region but also pay attention to the other boxer, which shows our CMA block can capture long-range dependencies. Moreover, not all the moving objects can attract the attention, only key information does. In the last example \textit{tying bow tie}, the block pays more attention to the region around the hand although the whole upper body is moving, because that the object held in hands often has more impact on the prediction. And more attention maps can be found in Appendix~\ref{sec:more_atm}.}
\label{fig:attentionmap}
\centering
\end{figure*}

\subsection{Comparison with Non-local Blocks}
\label{sec:vs_nl}

\begin{table}[thb]
\centering
\begin{small}
\begin{tabular}{ccccc}
\toprule
model & non-local & modality & top-1 & top-5 \\
\cmidrule(){1-5}
ResNet50 & - & RGB & 67.73 & 87.94 \\
ResNet50 & - & Flow & 55.73 & 79.04 \\
ResNet50 & - & RGB + Flow & 71.21 & 89.92 \\
\cmidrule(){1-5}
ResNet50 & Yes & RGB & 68.74 & 88.43 \\
ResNet50 & Yes & Flow & 56.66 & 80.11 \\
ResNet50 & Yes & RGB + Flow & 71.67 & 89.87 \\
\cmidrule(){1-5}
CMA\_$iter_1$-R & - & RGB + Flow & 72.17 & 90.70 \\
CMA\_$iter_1$-S & - & RGB + Flow & \textbf{72.62} & \textbf{91.04} \\
\cmidrule(){1-5}
CMA\_$iter_1$-R & Yes & RGB + Flow & 72.27 & 90.76 \\
CMA\_$iter_1$-S & Yes & RGB + Flow & 72.60 & 91.01 \\
\bottomrule
\end{tabular}
\end{small}
\caption{\small Comparisons with non-local networks on Kinetics.}
\label{tab:nonlocal}
\end{table}

Non-local block~\cite{NonLocal2018} is also a kind of attention-based model which pays attention to the intra-modality features. It also shows good performance in video classification. In order to compare the performance and mutual influence between self-attention blocks and our proposed cross-modality attention blocks, we carry out some experiments, and the results are shown in Table~\ref{tab:nonlocal}. Following~\cite{NonLocal2018}, we add five blocks to ResNet50 in $\text{res}_3$ and $\text{res}_4$, the same numbers and locations as that of the CMA blocks. To explore the influence between these two kinds of blocks, we conduct the experiments that adding CMA blocks just behind the nonlocal blocks. To ensure the comparisons more tractable, we only add the nonlocal blocks in the RGB branch, which implies that the Flow branch is the same to the Flow modality of the ResNet50 model.

From the results in Table~\ref{tab:nonlocal}, we find that non-local blocks can roughly improve top-1 accuracy by $1\%$ in both RGB and Flow modalities of ResNet50 model. For our proposed model, even only the results of the RGB branch outperform the fusion results of ResNet50 with nonlocal blocks. More importantly, the non-local blocks seem unnecessary while using our CMA blocks, which shows that the CMA blocks can also play a role of non-local blocks while fusing different modalities. In other words, the CMA blocks can also capture global information. We also visualize the attention maps of non-local blocks in Appendix~\ref{sec:more_atm}, which can intuitively show the improvement of CMA blocks over non-local blocks.

\subsection{3D-CMA Blocks}

\begin{table}[thb]
\centering
\begin{small}
\begin{tabular}{ccccc}
\toprule
{model} & \# input frames & RGB & Flow & two-stream \\
\cmidrule(){1-5}
P3D & 12 & 70.98 & 63.80 & 73.91 \\
P3D & 16 & 71.50 & 66.20 & 74.62 \\
\cmidrule(){1-5}
CMA\_${iter_1}$ & 12 & 74.41 & 63.80 & 75.22 \\
CMA\_${iter_1}$ & 16 & \textbf{74.86} & \textbf{66.20} & \textbf{75.98} \\
\bottomrule
\end{tabular}
\end{small}
\caption{\small Performance of P3D and 3D-CMA models on Kinetics when varying the count of input frames. All models adopt ResNet-152 as the backbone, and the input of CMA blocks are all 3D.}
\label{tab:3dcma}
\end{table}

To illustrate that the proposed CMA blocks can also be compatible with 3D convolutional neural networks and further improve its performance, we insert this operation into P3D ResNet~\cite{qiu2017P3D}. We initialize the P3D network with the weights duplicated from the official caffemodel\footnote{https://github.com/ZhaofanQiu/pseudo-3d-residual-networks} and finetune it using data argumentation. For the CMA model, considering the limited GPU memory, we only add one CMA block after the last layer in res$_4$, and train the CMA block and all layers behind it.
We train our CMA model with different numbers of input frames. Table~\ref{tab:3dcma} summarizes the experimental results. As seen, the CMA block can also bring an improvement for P3D compared with two-stream. Fusion with two branches can further improve the accuracy.

\subsection{Transfer learning}

\begin{table}[thb]
\centering
\begin{small}
\begin{tabular}{ccc}
\toprule
model & use 3D-Conv & top-1 \\
\cmidrule(){1-3}
ECO~\cite{Zolf2018ECO} & Yes & 94.8 \\
ARTNet~\cite{wang2017arnet} & Yes & 94.3 \\
I3D~\cite{CarreiraZ17i3d} & Yes & 98.0 \\
\cmidrule(){1-3}
Two-stream\cite{simonyan14two} & No & 88.0 \\
TSN~\cite{wang16tsn} & No & 94.0 \\
Attention Cluster~\cite{long2018attention} & No & 94.6 \\
\cmidrule(){1-3}
ResNet50-R & No & 90.9 \\
ResNet50-F & No & 92.4 \\
ResNet50-S & No & 95.5 \\
CMA\_$iter_1$-R & No & 95.3 \\
CMA\_$iter_1$-S & No & 96.5 \\
\bottomrule
\end{tabular}
\end{small}
\caption{\small Comparison with state-of-the-art on the UCF-101. The first set is the results reported by other papers, and the second set is our results of transfer learning.}
\label{tab:ucf101}
\vspace{-0.2in}
\end{table}

We also conduct transfer learning experiments from Kinetics to UCF-101. We only fine-tune the last fc layer of our 2-D CMA model. Table~\ref{tab:ucf101} shows the results. We find that our model is somewhat easier to over-fit on the small dataset. Nonetheless, the proposed CMA\_$iter_1$-S can outperform most of the state-of-the-art 2D models. It even approximates the performance of 3D models (\eg, I3D with 64 RGB frames and 64 flows as its input) although only 2D convolutional network as base model is used.

\section{Conclusion and Future work}

We have shown that the cross-modality operation can significantly improve the performance in video classification. The proposed CMA block can be compatibly inserted to most existing neural networks. It proves very effective to fuse information between different modalities. Our future works include extending the evaluations on other more sophisticated deep models, and evaluating the CMA operation among more modalities beyond both the RGB and Flow branches.


\begin{appendices}
\section{More attention maps}
\label{sec:more_atm}

To more intuitively illustrate the effect of CMA blocks, we show more attention maps of the last CMA block of CMA\_$iter_1$-R in Figure~\ref{fig:attentionmap}. Consistent with the conclusions in the main text, our CMA blocks tend to focus on moving objects (which are strong evidence for video classification), and are able to capture long-range dependencies. In experiments we also notice some failure cases. Figure~\ref{fig:attentionmap} highlights two examples using red bounding boxes. As seen, when the query position is located at the background, the CMA block may focus on itself or other positions of background, which supposedly does not benefit the performance.

\begin{figure*}[t]
\centering
\includegraphics[width=.9\linewidth]{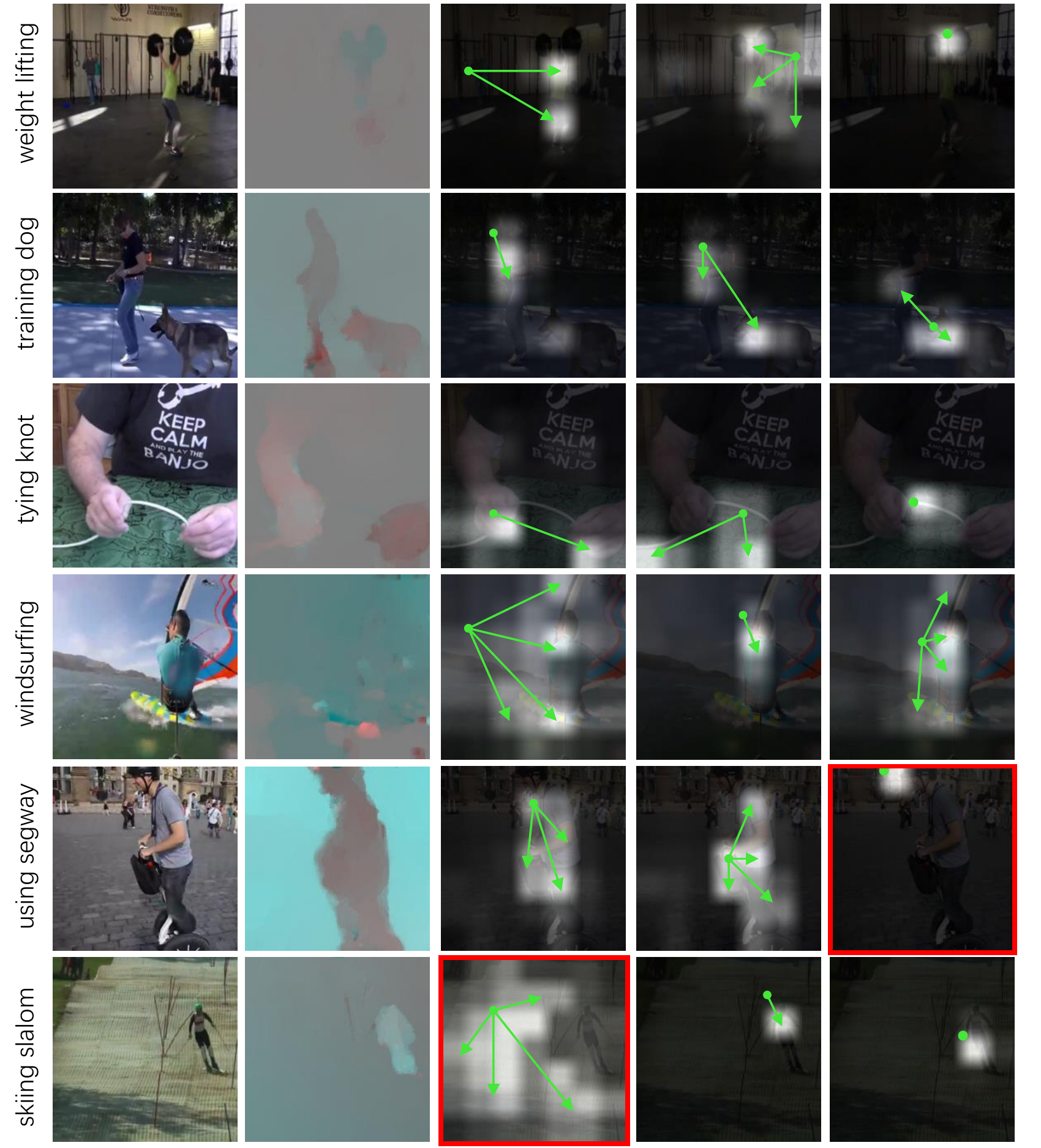}
\caption{\small \textbf{Examples of attention maps in CMA}. The columns from left to right are RGB frame, optical flow field and attention maps of three different interested positions. Examples highlighted by red bounding boxes are failure cases.}
\label{fig:attentionmap}
\centering
\end{figure*}

\begin{figure*}[t]
\centering
\includegraphics[width=.9\linewidth]{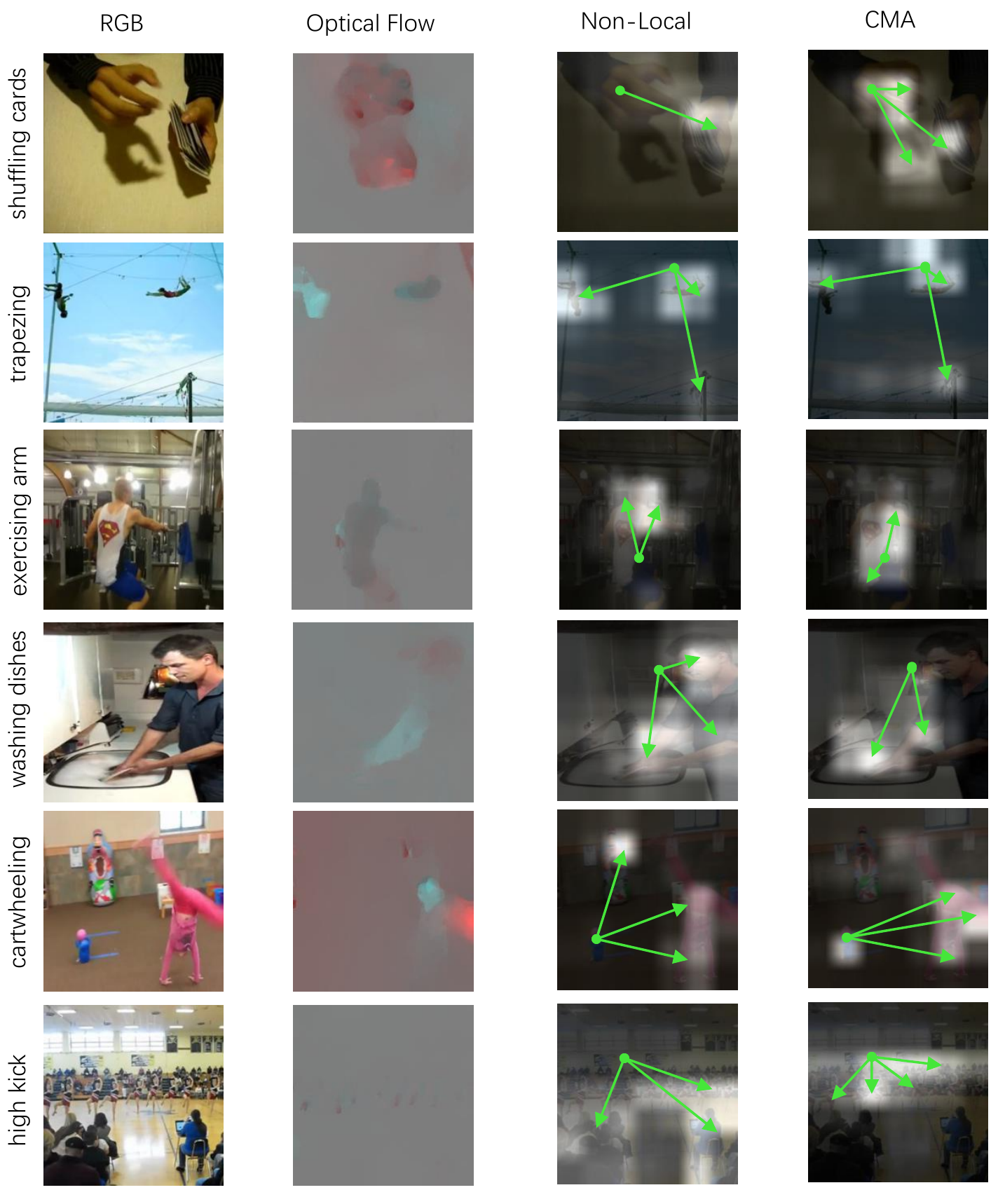}
\caption{\small \textbf{Examples of Non-local and CMA's attention maps}.}
\label{fig:cma-nl}
\centering
\end{figure*}
\end{appendices}

We also compare the attention maps generated by non-local block~\cite{NonLocal2018} and CMA block in Figure~\ref{fig:cma-nl}. The non-local blocks are specified exactly the same to the description in Section~\ref{sec:vs_nl} in the main text, adopting ResNet50 as backbone and using RGB frames as input. From Figure~\ref{fig:cma-nl}, we find that these two kinds of blocks have similar behavior in some examples where both focus on critical objects in the videos (like the one from action \emph{trapezing}). Nonetheless, our proposed CMA block can pay more attention to the moving objects which are at least equally crucial to the final prediction. For example, in the last two cases in Figure~\ref{fig:cma-nl}, representing actions of \textit{cartwheelin} and \textit{high kick} respectively, CMA block only focuses on the moving person while the non-local block disperses its attention to the stationary object or the audiences that are less relevant to video classification.

\end{document}